\documentclass[12pt,twoside]{article}

\usepackage[english]{babel}
\usepackage[utf8]{inputenc}

\usepackage{johd}          
\usepackage{amsmath,amssymb}
\usepackage{tikz}
\usetikzlibrary{automata,positioning,shapes.geometric,arrows}
\usepackage{caption}          
\usepackage{subcaption}       
\usepackage{apacite}
\usepackage{array}
\usepackage{rotating}
\usepackage{booktabs}
\usepackage{longtable}
\usepackage{tabularx}
\usepackage{multicol}
\usepackage{graphicx}
\usepackage{xifthen}
\usepackage{setspace}
\usepackage{chngpage}
\usepackage{csquotes}
\usepackage{fancyhdr}
\usepackage{geometry}
\usepackage{adjustbox}
\usepackage{pifont}
\usepackage{xcolor}
\usepackage{tcolorbox}
\tcbuselibrary{skins,breakable,listings,listingsutf8}
\tcbuselibrary{listingsutf8}
\usepackage{fvextra}
\usepackage{hyperref}
\usepackage{authblk}    
\usepackage{lscape}       
\usepackage{float}        

\tcbset{
  myprompt/.style={
    enhanced,
    breakable,
    colback=black!2,
    colframe=black!35,
    coltitle=black,
    fonttitle=\bfseries,
    boxrule=0.6pt,
    arc=2mm,
    left=8pt,right=8pt,top=8pt,bottom=8pt,
  }
}
\definecolor{codepurple}{RGB}{140,0,160}
\newtcblisting{codeblock}{
    colback=white,
    colframe=gray!60,
    listing only,
    listing options={
        basicstyle=\ttfamily\color{codepurple}\small,
        language=Python,
        breaklines=true,
        keywordstyle=\color{codepurple},
        showstringspaces=false
    },
    left=2mm,
    right=2mm,
    top=2mm,
    bottom=2mm,
    boxrule=0.5pt,
    arc=2pt,
    width=\textwidth
}

\tikzset{
  process/.style = {rectangle, draw, text centered, minimum width=4cm, minimum height=1cm},
  arrow/.style   = {thick, -{Stealth}}
}

\title{Beyond the Link: Assessing LLMs' Ability to Classify Political Content Across Global Media}

\author[1,2,$\dagger$]{Alejandro De La Fuente-Cuesta}
\author[1,3,$\dagger$]{Alberto Martinez-Serra}
\author[1]{Nienke Visscher}
\author[2]{Laia Castro}
\author[3,*]{Ana S.~Cardenal}

\affil[1]{Barcelona Supercomputing Center (BSC), Plaça Eusebi Güell 1-3, 08034 Barcelona, Spain}
\affil[2]{Department of Political Science, Universitat de Barcelona (UB), Avinguda Diagonal 684, 08034 Barcelona, Spain}
\affil[3]{School of Law and Political Science, Universitat Oberta de Catalunya (UOC), Rambla del Poblenou 154-156, 08018 Barcelona, Spain}

\affil[$\dagger$]{These authors contributed equally to this work}
\affil[*]{\textit{Corresponding author:} \texttt{acardenal@uoc.edu}}

\date{October 2025}

\begin{document}
\maketitle
\newpage

\begin{singlespace}
\begin{abstract}
The use of large language models (LLMs) is becoming common in political science and digital media research. While LLMs have demonstrated ability in labelling tasks, their effectiveness to classify Political Content (PC) from URLs remains underexplored. This article evaluates whether LLMs can accurately distinguish PC from non-PC using both the text and the URLs of news articles across five countries (France, Germany, Spain, the UK, and the US) and their different languages. Using cutting-edge models, we benchmark their performance against human-coded data to assess whether URL-level analysis can approximate full-text analysis. Our findings show that URLs embed relevant information and can serve as a scalable, cost-effective alternative to discern PC. However, we also uncover systematic biases: LLMs seem to overclassify centrist news as political, leading to false positives that may distort further analyses. We conclude by outlining methodological recommendations on the use of LLMs in political science research.
\end{abstract}
\end{singlespace}


\vspace{1cm}

\section{Introduction}\label{sec:Intro}

In the era of digital media, understanding how individuals consume information online has become central to the study of behavioural science \citep{martinez2025porn,kulshrestha2021characterizing}. 
The internet---and particularly news content distributed through websites and social media platforms---now serves as a primary source of political knowledge worldwide \citep{shehata2021learning}. 
This digital transformation has created new opportunities for researchers to investigate political information exposure at an unprecedented scale, using novel forms of behavioural data such as web tracking data. 
The use of large repositories of browsing histories can reveal de facto information diets of thousands of individuals \citep{christner2022automated,clemm2024analysis}. 
These datasets allow researchers to map news consumption patterns, such as the prevalence of selective exposure  and exposure to misinformation \citep{cardenal2019digital,  Wojcieszak2022Avenues,guess2020exposure}. 
However, researchers face a rather complex question: How can we reliably identify what counts as Political Content (PC)? 
In political communication studies, researchers typically classify news content based on whether the Uniform Resource Locator (URL) originates from a known media outlet. 
However, this approach usually struggles to indicate the content itself. Therefore, distinguishing PC from non-PC at large scale remains a significant challenge.

Large language models (LLMs) offer a promising new approach for scalable text classification in political analysis \citep{lemens2025positioning,kirkizh2024predicting}. 
Advances in Natural Language Processing (NLP) enable researchers to leverage the capacity of pretrained models for text classification \citep{wang2023pretrained,min2023recent}. 
Recent studies have shown that LLMs can serve as high-quality, cost-effective annotators for a variety of political text classification tasks, following the path of automated content analysis methods \citep{grimmer2013text}. 
For instance, \citet{heseltine2024large} demonstrated that GPT-4 can consistently replicate expert human coding for brief pieces of text from four nations. 
Similarly, \citet{vera2024bias} showed that LLMs reflect human-like biases in PC classification, particularly when party cues are present. 
Recently, \citet{lemens2025positioning} demonstrated that instruction-tuned LLMs can effectively position political texts on ideological scales.

Despite this progress, important gaps remain in our understanding of LLM-based classification, particularly for URL-only inputs and cross-lingual settings. Most prior studies have given LLMs access to the full text of documents (tweets, articles, etc.) when performing classification \citep{makhortykh2024panning}. 
An intermediate approach is to classify article titles, which were recently found to capture much of the context as proven by a high agreement between title- and full-text classification \citep{heseltine2025partisan}. 
However, in large-scale web-tracking, titles are rarely in raw trace data and require scraping each URL. Therefore, in many social media research scenarios where raw tracking data is provided, scholars possess only the URLs rather than the content itself \citep{olejnik2014uniqueness}.

As a result, a second question arises: Could LLMs accurately determine whether a URL links to PC without accessing the full article? 
If so, this would be a methodological advancement -- enabling rapid classification using minimal information. 
URL-based classification is not unprecedented in web research: previous approaches have used features like domain names, URL paths, or metadata to categorise webpages (for instance, to detect spam or phishing) with some success \citep{rastakhiz2024quickcharnet,li2025continuous,baykan2013comprehensive}.

With news, the URL usually contains cues in the path -- for example, keywords from the article title -- that an advanced LLM might be able to use \citep{clemm2024analysis}. 
However, some outlets rely on highly non-descriptive URL structures, providing little linguistic information that models can exploit (e.g., /world-europe-60547473). 
This further complicates URL-based classification, since such cases offer almost no textual signal about the underlying content. 
Nevertheless, the potential advantages of URL-only classification are significant: It eliminates the need for fetching or processing full text, speeding up analysis and avoiding copyright or paywall issues. 
Recent evidence indicates that in situations where full content is inaccessible, classification performance can still be sufficient using solely URLs \citep{schelb2024assessing}.

In this study we investigate this question by evaluating several leading LLMs on the task of classifying political versus non-political news, evaluating both URLs and full text scenarios. 
We present, to our knowledge, the first LLM assessment on PC classification across multiple countries and languages. 
We test a range of state-of-the-art models, such as Llama \citep{grattafiori2024llama}, Mistral \citep{jiang2024mixtral}, DeepSeek \citep{deepseek2025deepseek}, Qwen \citep{yang2025qwen3}, and Gemma \citep{gemma2025technical}, to examine how different models and sizes impact performance (Table~\ref{tab:llm-summary}).

By comparing model predictions to a human-labelled ground, we assess the accuracy of each LLM in identifying PC. 
Our results show that LLMs can classify PC with high accuracy not only from full articles but also from URLs alone, with URL-based approaches often improving precision while remaining consistent across countries and languages. 
However, we also find that LLMs tend to overestimate the presence of PC in centrist news, leading to systematic false positives that may bias estimates of political exposure and polarisation. 
Overall, this work extends the current methods of content analysis, and our descriptive knowledge of the ways people access news in the digital age, in addition to practical recommendations on LLM-based methodologies for future research in political analysis.

\section{Data and Methods}\label{sec:Data&Methods}

\subsection{LLMs}\label{sec:Data&Methods}

To classify the PC of the URLs and databases, we made use of different large language models (LLMs) listed in Table \ref{tab:llm-summary}. This selection includes models of varying sizes and providers, ranging from computationally demanding architectures to smaller ones like Llama-8B that can be more easily run locally. All experiments were conducted on a high-performance computing facility available at the authors’ institution \footnote{As reported in the Supplementary Information, we additionally ran the same experiments with ChatGPT-4.5. Results were substantively consistent with those presented here, confirming that the findings do not depend on the use of commercial models.}. The zero-shot prompt used can be seen in Figure~\ref{fig:classification-prompt}. In addition to the main prompt, we also implemented an abstention option (‘SKIP’) for cases in which the URL path was empty, encoded, or contained no meaningful linguistic cues. This design avoids forcing arbitrary classifications that could otherwise introduce noise into the results. Details and results for the SKIP condition are reported in the Supplementary Information.

\begin{table}[ht]
\begin{adjustwidth}{-0.0cm}{-0.0cm} 
\centering
\footnotesize
\renewcommand{\arraystretch}{1.0}
\setlength{\tabcolsep}{5pt}
\begin{tabularx}{\linewidth}{@{} l X c c l @{}}
\toprule
\textbf{LLM} & \textbf{Full Name} & \textbf{Execution} & \textbf{End of Training} & \textbf{Publisher} \\
\midrule
DeepSeek R1 7B     & R1-Distill-Qwen-7B               & Open (Local) & Apr 2025 & DeepSeek AI \\
Gemma 3 27B        & gemma-3-27b-it                   & Open (Local) & Aug 2024 & Google DeepMind \\
Llama 3.1 8B       & Llama-3.1-8B-Instruct            & Open (Local) & Dec 2023 & Meta \\
Mistral Small 2IB  & Mistral-Small-3.1-2IB-Instruct-2503 & Open (Local) & Mar 2024 & Mistral AI \\
QwQ 32B            & QwQ-32B                          & Open (Local) & Mar 2025 & Alibaba Cloud \\
\bottomrule
\end{tabularx}
\caption{\textbf{Summary of evaluated LLMs}, including model names, execution types, training completion dates, and publishers.}
\label{tab:llm-summary}
\end{adjustwidth}
\end{table}

\begin{minipage}{0.98\linewidth}
  \begin{tcolorbox}[myprompt,title={Classification Prompt},breakable]
    \ttfamily
    \scriptsize
    \begin{Verbatim}[breaklines,breakanywhere]
analyse the following text to identify:
    1. Whether it describes a political EVENT or action. Political content means explicit reference to:
       - Governments, parliaments, elections, parties, politicians, political leaders, public policies, laws, state institutions, or protests/mobilisations with political actors.
       - Decisions, statements, or conflicts involving political institutions.
       Do NOT classify as political if the text is only about social, cultural, scientific, or economic issues without reference to political actors or decision-making.
    2. Political position on a 1-10 scale (1 = left, 10 = right)
    **Paragraph:**
    "{paragraph}"
    **Required JSON Format:**
    {{"Answer": "Yes" or "No",
      "PoliticalPosition": number (1-10) or null}}
    **Instructions:**
    - "Answer" = "Yes" ONLY if the text refers to political events, decisions, or statements involving governments, parties, politicians, institutions, or elections.
    - If the text mentions broad social topics (climate change, gender, health, etc.) WITHOUT political actors, classify as "No".
    - "PoliticalPosition": use 1-3 for left, 4-6 for center, 7-10 for right. Use null if "Answer"="No".
    - Use exactly this JSON format, no markdown, no extra text.

    **Special Notes:**
    - Identify ideological markers only when linked to political actors.
    - Exclude purely cultural, scientific, or lifestyle news without political context.
    Return ONLY valid JSON.
    \end{Verbatim}
  \end{tcolorbox}
  \captionof{figure}[Classification prompt used in all the models]{Classification prompt used in all the models.}
  \label{fig:classification-prompt}
\end{minipage}
\vspace{10pt}

The code structure was adapted according to the reasoning of each model, as well as for the text-only scenario \citep{tornberg2024best}. To classify the URLs from the dataset, we first filtered the visits from a web tracking sample by those containing a web domain related to political news \citep{DVN/NGWUB7_2025}. All pre-selected URLs were web-scrapped and later manually classified by two experts into “POL” (political) and “NON” (non-political) based on their content. In those texts with disagreement, a third expert assigned a final category. Broadly, we classified all hard news articles covering local, regional, national, and international politics as POL, and everything else as NON. However, the line separating “POL” from “NON” was not always easy to draw.  For each country, we classified as many articles as necessary to have $\sim$200 political and $\sim$200 non-political human-coded news articles per country. To assess the reliability of the human annotations, we computed intercoder agreement between the two coders. The observed agreement was 96.6\%, and Cohen’s kappa reached 0.93 ($z = 30.9$, $p < 0.001$), indicating almost perfect agreement \citep{landis1977measurement}. Data was accessible for 81\% of these files but 18\% returned no data because of moved URLs and other accessibility problems \citep{christner2022automated,adam2024improving}.

\subsection{Data}

This study used real data collected from a web tracking across five countries: France (FR), Germany (DE), Spain (ES), the United Kingdom (UK) and the United States (US). The web tracking data recorded all websites visited via laptops/desktops, mobile phones and tablets, including timestamps, session duration, and total time spent on each site.

Respondents were recruited from an invitation-only online survey panel managed by a commercial provider, which ensures diverse socio-demographic representation per country through a controlled recruitment process. Panellists were selected through a double opt-in procedure to prevent duplication and maintain data integrity. The experiment registered the online activity of the panellists from February 22nd to June 5th, 2022.

To study the categorisation of news as political or non-political, the web tracking dataset was filtered to focus on visits to selected media websites. The lists of media outlets were compiled using ComScore data on the top 50-100 most popular news outlets by country. We completed it by adding news outlets that were among the top 1,000 most visited websites in our country samples. The data was aggregated at individual level, and 1,140 visits were randomly chosen out of the total, an amount compatible with human categorisation. This human categorisation was done through a two-step blind coding method. First, two coders marked all the 1,140 randomly sampled visits separately without reference to any identifying information concerning its previous categorisation in an attempt to eliminate biased judgment. Coders marked each content as POL (it has PC) or NON (It does not have PC). In the second step, conflicts among the categorisation made by the first coders were raised and checked by a third, independent coder who served as an adjudicator. This adjudicated outcome was used as the gold standard (‘Human’) reference in all main analyses. 

\section{Results}

To evaluate the effectiveness of LLMs in classifying political versus non-political news content, we analyse two modalities: full article text and URL-based inputs. Performance is assessed across multiple LLMs against a human-coded annotations baseline.

\begin{figure}[htbp]
  \centering
  \includegraphics[width=\linewidth]{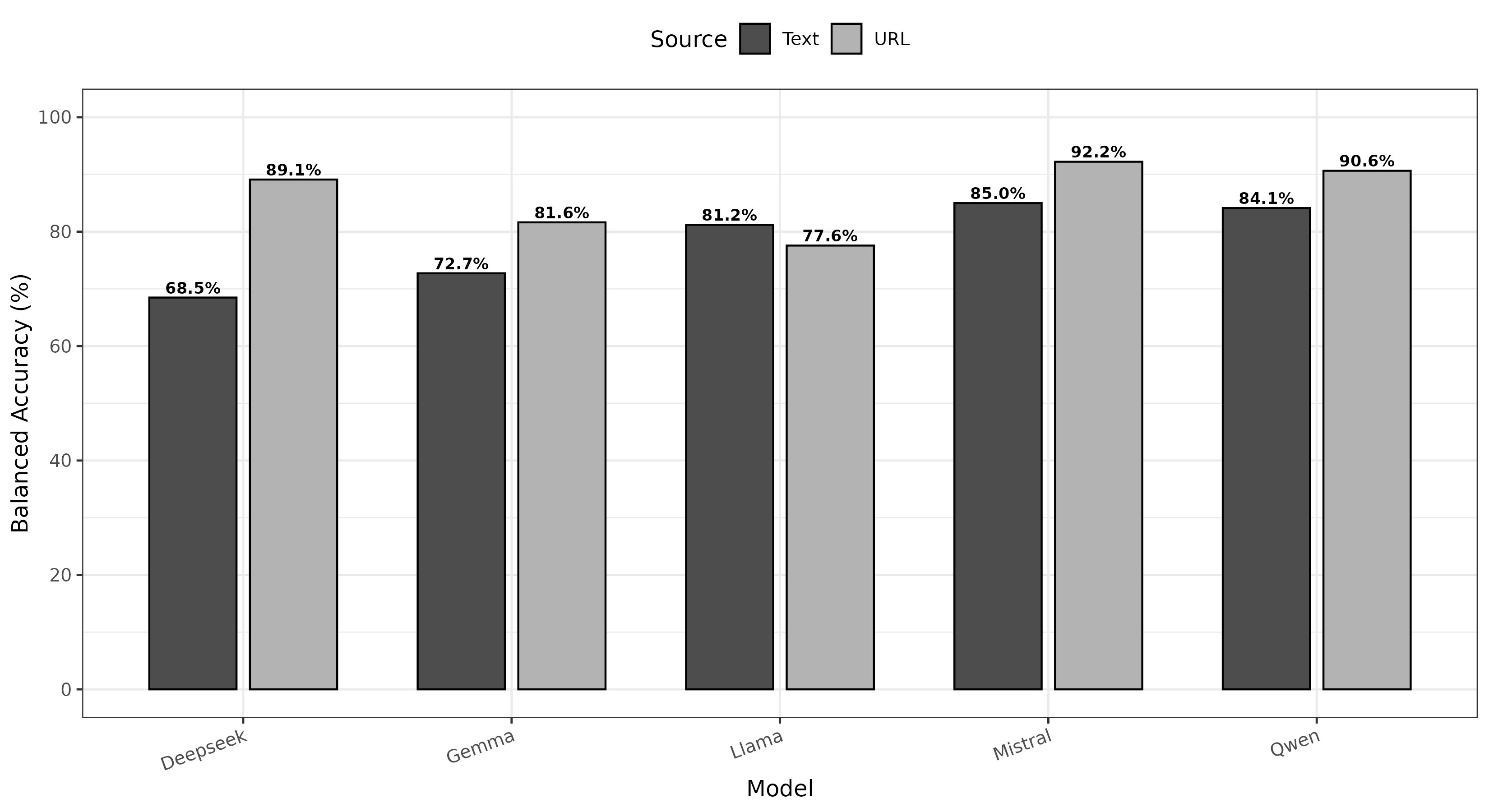}
  \caption{Balanced Accuracy by model, using human classification as the reference.}
  \label{fig:balanced_accuracy_human}
\end{figure}

Figure \ref{fig:balanced_accuracy_human} shows the balanced accuracy, understood as the average of sensitivity (the share of political items correctly identified) and specificity (the share of non-political items correctly identified). The results indicate that models generally achieve higher balanced accuracy when classifying based on URLs than when using the article text. For example, Mistral and Qwen reach over 90\% balanced accuracy with URLs, while their text-based performance is closer to 85\%. Deepseek shows the largest gap (68.5\% with text vs. 89.1\% with URLs), suggesting it relies more heavily on domain cues. Overall, the findings highlight that URL information alone is often sufficient, and in some cases even superior, to identify whether content is political.

\begin{figure}[htbp]
  \centering
  \includegraphics[width=\linewidth]{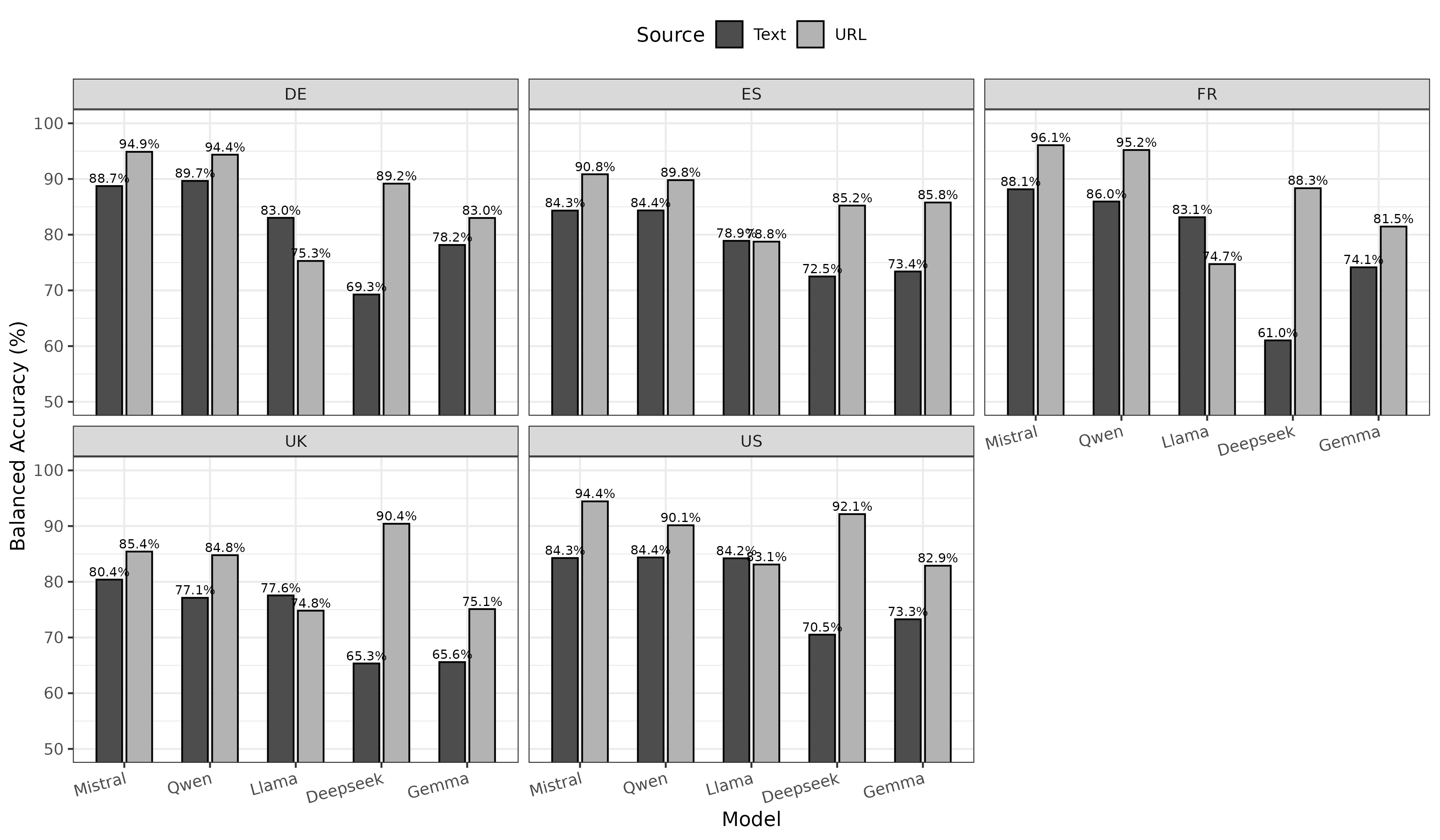}
  \caption{Balanced accuracy by model across countries, using human classification as a reference.}
  \label{fig:balanced_accuracy_country}
\end{figure}

Moreover, we examined cross-country variation in model–human agreement, aggregating results across all evaluated models. This view allows us to assess whether the PC classification is consistent across different national contexts. As shown in Figure \ref{fig:balanced_accuracy_country}, results indicate that the balances agreement with human annotations remains high in all countries, with minimal and non-significant differences, suggesting that linguistic, cultural, or media system factors may residually influence model performance for both text and URL classification.

\begin{figure}[htbp]
  \centering
  \includegraphics[width=\linewidth]{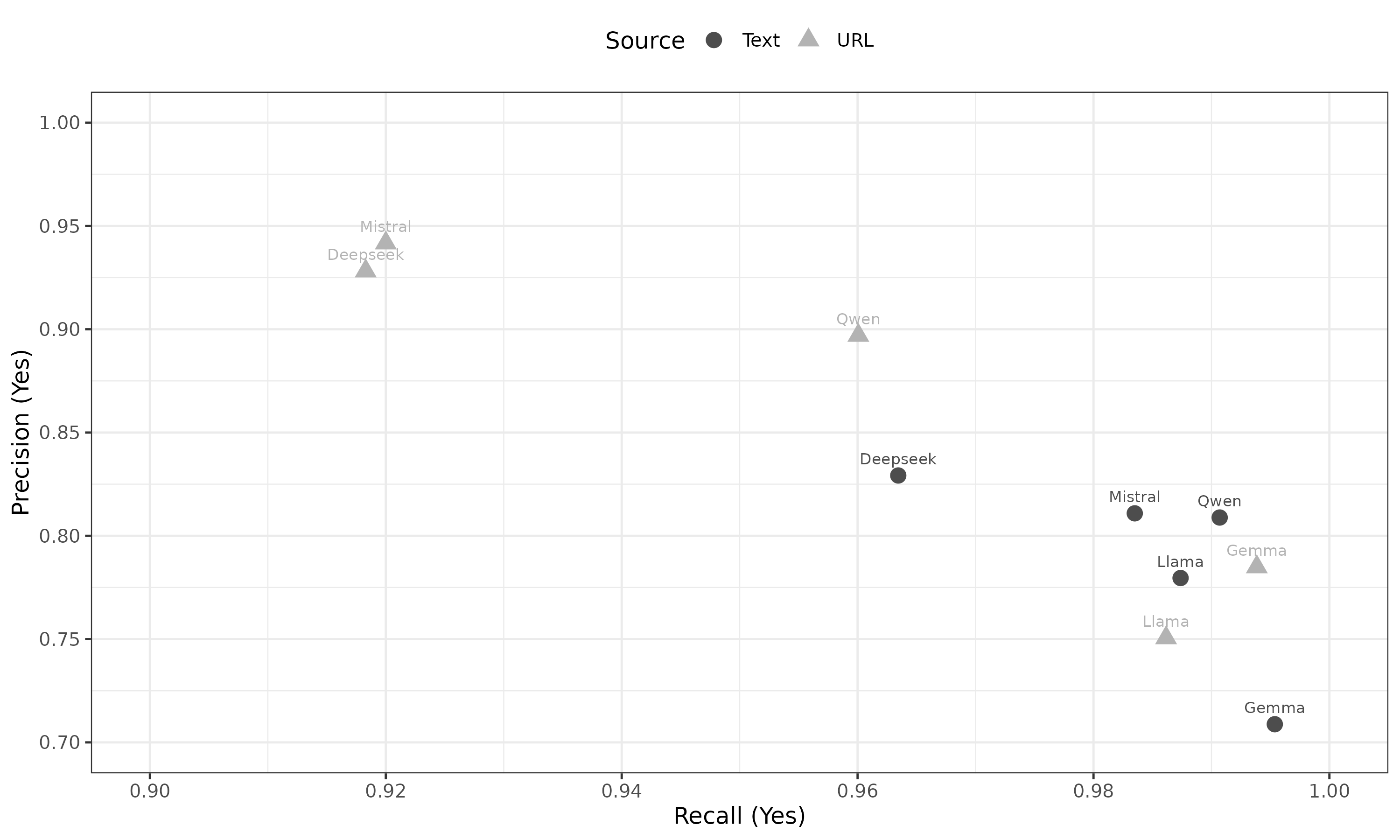}
  \caption{Precision versus Recall for the positive class (Yes), comparing Text and URL predictions against human classification.}
  \label{fig:precision_recall_human}
\end{figure}

Figure \ref{fig:precision_recall_human} illustrates the trade-off between precision and recall for the political class. Models that rely on text consistently appear in the lower right of the plot, achieving very high recall (0.97–0.99) and capturing almost all political articles, but at the cost of lower precision, that is, generating a considerable number of false positives (with precision ranging from 0.71 to 0.83). By contrast, models that rely on URLs cluster in the upper left of the plot, with slightly lower recall (0.92–0.94) but higher precision (0.90–0.95). In practical terms, URL-based classifications are more conservative: they identify PC less frequently, but when they do, they are more likely to be correct. Overall, the pattern suggests that URL signals substantially improve precision without a harmful reduction in recall, raising the F1 score for the positive class to 93\% (Table \ref{tab:models_performance}).

Cohen’s Kappa metric corroborates the F1 story and provides a chance-adjusted metric of reliability. Most URL settings reach substantial to almost perfect agreement: Mistral–URL $\kappa$=0.84 and Qwen–URL $\kappa$=0.82, with Deepseek–URL $\kappa$=0.78 and Gemma–URL $\kappa$=0.66 having a substantial correlation. Text variants are consistently lower (e.g., Mistral–Text $\kappa$=0.72, Qwen–Text $\kappa$=0.71, Deepseek–Text $\kappa$=0.44, Gemma–Text $\kappa$=0.49), indicating more disagreement beyond chance when relying on content alone.

\begin{table}[htbp]
\begin{adjustwidth}{-0.75cm}{-0.75cm} 
\centering
\footnotesize
\label{tab:models_performance}
\renewcommand{\arraystretch}{1.0}
\setlength{\tabcolsep}{2pt}
\begin{tabularx}{\linewidth}{l l c c c c c c}
\toprule
\textbf{Model} & \textbf{Source} & \textbf{Accuracy} & \textbf{Balanced Acc.} & \textbf{F1 (Yes)} & \textbf{Specificity} & \textbf{MCC} & \textbf{Kappa} \\
\midrule
Deepseek & Text   & 82.4\% [79.9–85.1\%] & 68.5\% & 89.1\% [87.4–90.9\%] & 40.6\% & 0.48 & 0.44 \\
Deepseek & URL    & 90.0\% [88.5–91.6\%] & 89.1\% & 92.3\% [91.0–93.6\%] & 86.4\% & 0.78 & 0.78 \\
Gemma    & Text   & 76.4\% [74.7–78.3\%] & 72.7\% & 82.8\% [81.2–84.3\%] & 45.9\% & 0.56 & 0.49 \\
Gemma    & URL    & 84.1\% [82.4–85.7\%] & 81.6\% & 87.7\% [86.3–89.0\%] & 63.9\% & 0.70 & 0.66 \\
Llama    & Text   & 83.5\% [81.9–85.2\%] & 81.2\% & 87.1\% [85.7–88.7\%] & 63.7\% & 0.69 & 0.65 \\
Llama    & URL    & 80.5\% [78.8–82.2\%] & 77.6\% & 85.2\% [83.8–86.6\%] & 56.5\% & 0.63 & 0.58 \\
Mistral  & Text   & 86.4\% [84.9–87.9\%] & 85.0\% & 88.9\% [87.5–90.2\%] & 71.6\% & 0.74 & 0.72 \\
Mistral  & URL    & 92.2\% [91.0–93.3\%] & 92.2\% & 93.1\% [92.0–94.1\%] & 92.4\% & 0.84 & 0.84 \\
Qwen     & Text   & 86.2\% [84.7–87.9\%] & 84.1\% & 89.1\% [87.8–90.4\%] & 69.2\% & 0.73 & 0.71 \\
Qwen     & URL    & 91.4\% [90.2–92.7\%] & 90.6\% & 92.7\% [91.7–93.9\%] & 85.3\% & 0.83 & 0.82 \\
\bottomrule
\end{tabularx}
\caption{Model performance metrics (reference: Human Classification)}
\label{tab:models_performance}
\end{adjustwidth}
\end{table}

\begin{figure}[htbp]
  \centering
  \includegraphics[width=\linewidth]{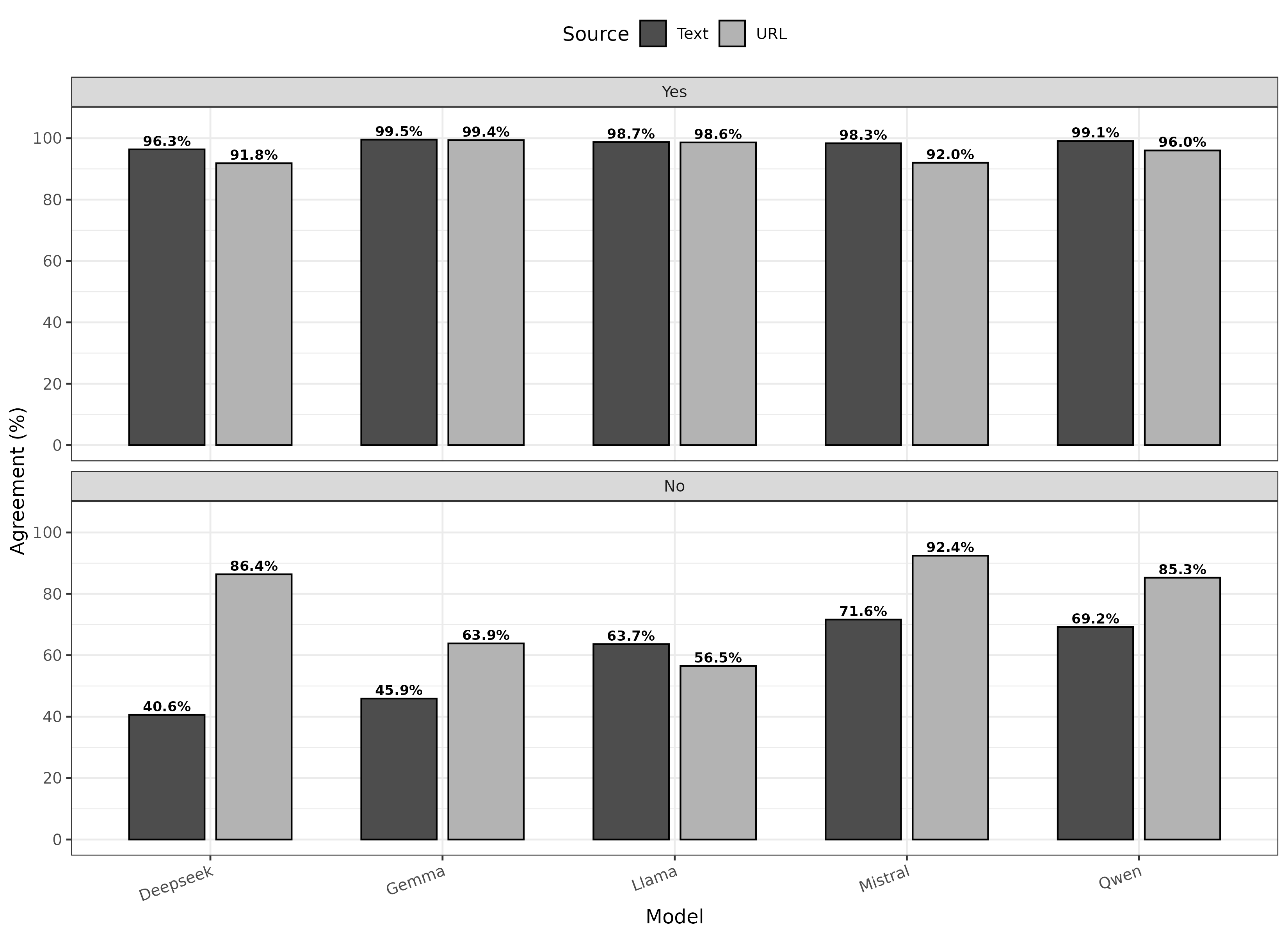}
  \caption{Agreement with human classification by class (Yes/No) across models and sources.}
  \label{fig:agreement_class_human}
\end{figure}

To conduct an in-depth analysis and avoid biases when using the automatic classification of LLMs, it is necessary to examine the sources of disagreement with human coding. Once agreement is analysed by class (Figure \ref{fig:agreement_class_human}), it becomes evident that LLMs are able to identify the majority of cases in which the true label is “Yes” (ranging from 91.8\% to 99.5\%), whereas their agreement drops substantially when the true label is negative (40.6\% to 92.4\%). These results suggest that LLMs exhibit a bias toward false positives, labelling as PC items that human annotators did not interpret as such.

\begin{figure}[htbp]
  \centering
  \includegraphics[width=\linewidth]{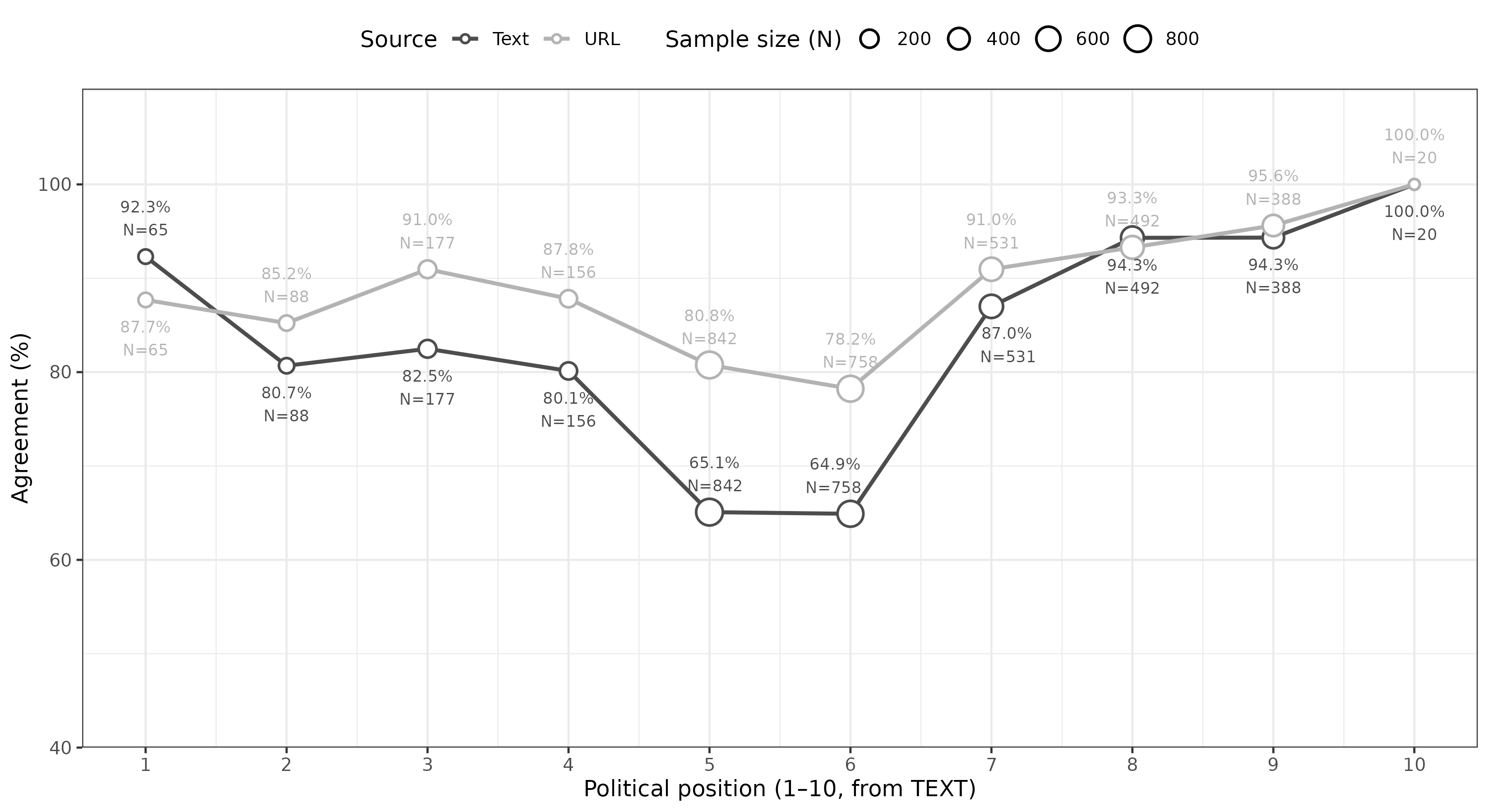}
  \caption{Weighted overall agreement by political position (1–10), comparing Text and URL predictions with human classification.}
  \label{fig:agreement_position_human}
\end{figure}

Therefore, it is necessary to investigate further where this bias originates. To this end, the language models were asked to provide an assessment of the political ideology of the text on a 1–10 scale. While we do not independently validate this ideological classification (see \citet{lemens2025positioning}, for a systematic evaluation of LLM-based ideology placement), our goal here is not to measure ideology but to use the relative scale as a diagnostic tool to identify where model–human disagreements concentrate. For this purpose, even if the ideological labels are noisy, they suffice to reveal whether misclassifications are disproportionately clustered in certain parts of the ideological spectrum. Once this classification was obtained, and after ordering the articles according to the classification based on the full-text context, an agreement analysis was conducted. The results are shown in Figure \ref{fig:agreement_position_human}. As can be observed, agreement decreases substantially in the central values of the political spectrum (4–6), dropping from values close to 100\% for more extremist texts to around 65\% for texts with a less extreme political ideology.

\begin{figure}[htbp]
  \centering
  \includegraphics[width=\linewidth]{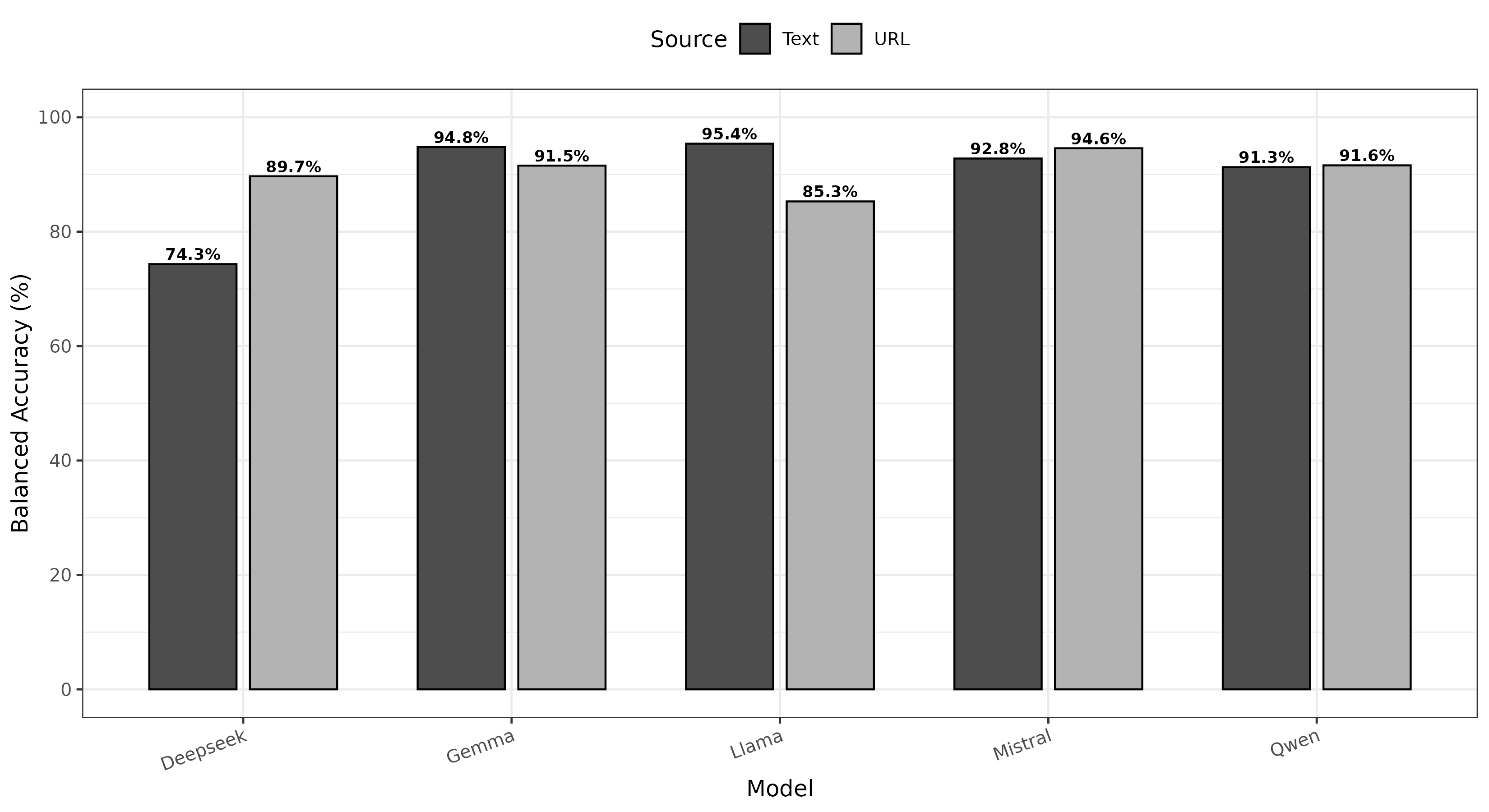}
  \caption{Balanced agreement by model and source, excluding items with political position in Text between 4 and 6.}
  \label{fig:balanced_agreement_filtered}
\end{figure}

In fact, once these news items are removed, the balanced agreement of the models increases systematically (Figure \ref{fig:balanced_agreement_filtered}). Beyond this improvement, the differences between the use of URLs and text decrease, with some cases in which text-based classification even outperforms URL-only classification (Gemma and Llama).

As observed in our specific case, LLMs tend to generate false positives, particularly for texts located at the political center. This suggests a potential bias in the results, which appears to be inherent not in the proposed methodology but in the models available at the current stage of development.

\section{Discussion}

The findings reported in this study highlight that LLMs are highly effective at classifying PC when provided with either full article text or just URLs. Most models up to date (April 2025) show strong alignment with human annotations when identifying PC in the text of articles. Notably, all the models demonstrate that even URL-only inputs can yield high-quality classifications, even better than using the whole text, offering a fast and scalable alternative to full-text analysis. This efficiency makes URL-based classification particularly attractive in large-scale research settings, where millions of records may be processed and full content is often inaccessible due to technical, legal, or privacy constraints.

However, model performance becomes more variable when limited to URL-only inputs. The success of URL-based classification depends not only on the model but also on how publishers structure their URLs. Some news outlets do not follow consistent or descriptive labelling conventions \citep{google2025url}, limiting the semantic cues available for LLMs to process. This issue adds noise to the classification task and poses challenges for generalisability across domains and countries. One way to mitigate such biases is to allow abstention when URLs provide no useful cues. In our supplementary analyses we operationalize this through a SKIP option, which reduces the influence of noisy or uninformative URLs. Importantly, results remain substantively consistent with those reported in the main text, reinforcing the robustness of our findings.

It is also important to consider the practical accessibility of the described method. While smaller models —particularly Llama 8B adn Deepseek R1 7B— demonstrated lower accuracy, this does not limit the utility of the method. Several mid-sized open-source models (e.g., Mistral, Gemma) achieved comparatively strong results, indicating that researchers with limited computational resources or restricted API access can still obtain valuable classifications. Consequently, the applicability of this approach extends beyond teams with high-powered infrastructure, though model choice should remain a deliberate methodological consideration. Moreover, its application extends beyond the context of web-tracking data. This approach may be particularly useful in other domains where large content is systematically hard to retrieve. For example, social media analysis is often based solely on URLs in their metadata streams, with no guarantee of stable or accessible full content. In such settings, URL-only classification could provide unique value, allowing researchers to recover meaningful information about the visited content.

Our findings also show that some models produce relatively few false negatives—correctly identifying most PC —but are more prone to false positives, misclassifying some non-PC as PC. This trade-off is important in studies where political knowledge (PK) is the outcome of interest. For instance, if researchers aim to assess whether exposure to PC increases PK, a high rate of false positives may introduce noise, potentially attenuating the estimated effects of political exposure. Conversely, if the goal is to examine the influence of non-PC on PK, minimizing false negatives becomes critical, as it ensures that non-PC exposure is not mistakenly counted as PC. These distinctions highlight the need for researchers to align model choice and classification thresholds with their specific research objectives.

Furthermore, researchers must take into account the type of bias that LLMs may introduce. Depending on the task at hand, different LLMs, due to differences in training, may yield different results for the same classification problem. In addition to always corroborating with human annotations as common ground and gold standard, it is crucial to examine in which subsamples the classifications diverge. As demonstrated in the results section, most of the bias stems from news considered to be at the political center. In our case, the ideological scores are used only as a diagnostic tool to identify where misclassifications occur, not as a substantive measure of ideology. However, if a researcher were interested in measuring the ideological orientation of news content, this bias could have different implications. In particular, the systematic overclassification of centrist items as political could pose problems for studies seeking to estimate polarisation based on news content, since it would artificially inflate the amount of PC attributed to centrist outlets. This compression of the ideological spectrum would reduce the apparent distance between centrist and partisan outlets, ultimately leading to an underestimation of polarisation.

Overall, URL-only classification offers a scalable, low-cost, and effective alternative for PC detection, especially when using top-performing models. We believe that these results will provide practical guidance for researchers seeking to use LLMs in large-scale media analysis where full content access is limited.

\section*{Declarations}

\textbf{Availability of data and material:} All data and replication materials are stored in a trusted open-data repository and will be released under a permanent DOI once the paper is accepted for publication.

\textbf{Ethical Approval:} The current study used data originally collected for research on political trends in France, Germany, Spain, the United Kingdom and the United States. All procedures involving human participants adhered to institutional and national ethical standards, the 1964 Helsinki Declaration and its amendments, and the ICC/ESOMAR International Code of Marketing and Social Research Practice.

\textbf{Informed Consent:} Participants were required to be legally competent and provided informed consent in accordance with EU-GDPR, allowing their data to be anonymously collected, purchased, and used for market and social science research.

\textbf{Conflicts of interest:} All authors certify that they have no affiliations with or involvement in any organisation or entity with any financial interest or non-financial interest in the subject matter or materials discussed in this manuscript.

\bibliographystyle{apacite}
\bibliography{bib.bib}

\end{document}